\documentclass[conference]{IEEEtran}
\IEEEoverridecommandlockouts

\usepackage{amsmath,amssymb,amsfonts}
\usepackage{algorithmic}
\usepackage{graphicx}
\usepackage{textcomp}
\usepackage{xcolor}
\def\BibTeX{{\rm B\kern-.05em{\sc i\kern-.025em b}\kern-.08em
    T\kern-.1667em\lower.7ex\hbox{E}\kern-.125emX}}

\usepackage{tikz}
\newcommand{\blackcircle}[1]{%
    \tikz[baseline=(char.base)]{
        \node[shape=circle,draw,fill=black,text=white,inner sep=1pt] (char) {#1};
    }%
}

\usepackage{balance}

\begin{document}

\title{Lesion-DDPM: Lesion-Enhanced 3D Diffusion for MS MRI Synthesis}

\author{\IEEEauthorblockN{Weidong Zhang\IEEEauthorrefmark{1}, Yongchan Jung\IEEEauthorrefmark{2}, Shafayat Mowla Anik\IEEEauthorrefmark{3}, Furen Xiao\IEEEauthorrefmark{4}, \\
Vasudevan Janarthanan\IEEEauthorrefmark{2},
Enkhzaya Chuluunbaatar\IEEEauthorrefmark{5}, Marco Ho\IEEEauthorrefmark{6}, Byeong Kil Lee\IEEEauthorrefmark{3}, and Jeeho Ryoo\IEEEauthorrefmark{2}}
\IEEEauthorblockA{\IEEEauthorrefmark{1}Northeastern University, Vancouver, BC}
\IEEEauthorblockA{\IEEEauthorrefmark{2}Fairleigh Dickinson University, Vancouver, BC}
\IEEEauthorblockA{\IEEEauthorrefmark{3}University of Colorado at Colorado Springs, Colorado Springs, CO}
\IEEEauthorblockA{\IEEEauthorrefmark{4}National Taiwan University, Taipei, Taiwan}
\IEEEauthorblockA{\IEEEauthorrefmark{5}University of British Columbia, Vancouver, BC}
\IEEEauthorblockA{\IEEEauthorrefmark{6}British Columbia Institute of Technology, Burnaby, BC\\
Email: zhang.weid@northeastern.edu, y.jung@student.fdu.edu, sanik@uccs.edu, \\
fxiao@ntu.edu.tw, v\_janart@fdu.edu, enkhzaya.chuluunbaatar@ubc.ca, marco\_ho@bcit.ca, blee@uccs.edu, j.ryoo@fdu.edu}
}

\maketitle

\begin{abstract}

3D FLAIR MRI is widely recommended as one of the standard MRI sequences for brain imaging in multiple sclerosis (MS), but publicly available MS datasets remain relatively small and vary across scanners, acquisition protocols, and lesion patterns. This scarcity and variability hinder the development of robust neuroimaging machine learning models and are particularly challenging for generative models that aim to synthesize images while preserving small, sparse lesions. We propose Lesion-DDPM, a 3D conditional diffusion framework for lesion-aware FLAIR synthesis that incorporates multi-level anatomical mask injection together with a lesion-weighted reconstruction loss to emphasize lesion voxels while maintaining global brain structure. Using a curated subset of the MSLesSeg dataset, we compare Lesion-DDPM with representative state-of-the-art GAN- and diffusion-based models, assessing both image-generation metrics and downstream 3D U-Net segmentation. In our experiments, Lesion-DDPM achieved the lowest lesion-region reconstruction error among all methods. In a downstream 3D U-Net lesion segmentation task, a model trained only on Lesion-DDPM–generated scans and evaluated on real MRIs reached a Dice score of 0.616 compared with 0.569 for the best competing synthetic dataset. When Lesion-DDPM images were added to the real training set, the Dice score further increased to 0.685.
\end{abstract}

\section{Introduction}
\label{sec:introduction}

Clinical brain MRI for multiple sclerosis (MS) is typically performed using multisequence protocols, but contemporary guidelines emphasize sagittal 3D FLAIR (fluid-attenuated inversion recovery) as a core sequence for diagnosis and monitoring due to its high sensitivity to supratentorial lesions and new lesion activity \cite{wattjes20212021}. By suppressing cerebrospinal fluid (CSF) signal, FLAIR improves the conspicuity of periventricular and juxtacortical hyperintensities—two hallmark lesion locations in MS—and can enhance visual detection compared with conventional T2-weighted imaging in these regions \cite{bakshi2001fluid}. Moreover, 3D FLAIR generally yields higher lesion counts and better contrast-to-noise ratios than 2D FLAIR, with multiple studies reporting substantially higher lesion detection across both supratentorial and infratentorial compartments \cite{bink2006detection,wang2018comparing}. While standard protocols include T2-weighted and T1-weighted sequences for comprehensive assessment, we focus on brain 3D FLAIR because it is widely treated as the primary sequence for maximizing lesion conspicuity around CSF interfaces and tracking new brain lesions in routine follow-up \cite{wattjes20212021}.

Motivated by its clinical role, we target 3D FLAIR as the synthesis objective and aim to generate high-fidelity synthetic volumes that preserve small MS lesions. Such synthetic data can mitigate data scarcity for model development and evaluation, yet public MS MRI resources remain comparatively small and heterogeneous. For example, the MICCAI MSSEG dataset contains only 53 patients scanned on multiple scanners with expert FLAIR/T2 annotations, while the newer MSSEG-2 challenge provides 100 patients split across training and test cohorts \cite{commowick2021multiple,kamraoui2022longitudinal}.

MS lesions are small, sparse, and morphologically irregular; in voxel space they occupy only a tiny fraction of brain volume, leading to severe lesion–background imbalance. These characteristics, combined with limited training data, make lesion visibility difficult to preserve: generative models may oversmooth subtle lesions or introduce artifacts. Prior work on conditional MRI synthesis includes GAN-based translation methods \cite{NIPS2014_Goodfellow, GANKwon, HierarchicalAmortizedGAN}, which are often challenged by training instability and mode collapse in biomedical settings \cite{saad2024survey}, diffusion-based 3D semantic synthesis with improved stability \cite{kazerouni2023diffusion}, and lesion compositing strategies for augmentation that still depend on real background brains and therefore are not fully synthetic \cite{Huo2024LesionGen}. This motivates a lesion-aware generative framework that maintains whole-brain realism while explicitly prioritizing subtle lesion fidelity.


We present Lesion-DDPM, a 3D conditional diffusion model for lesion-aware FLAIR synthesis with two key contributions. First, we employ multi-level conditional injection by introducing anatomical mask information at the encoder, bottleneck, and decoder stages, reinforcing lesion context across multiple feature scales and improving conditioning stability. Second, we incorporate a lesion-weighted L1 reconstruction objective that increases gradient emphasis on lesion voxels while maintaining overall image fidelity in non-lesion regions.

\section{Related Works}
Generative approaches for medical MRI synthesis can be broadly grouped into GAN-based image-to-image translation, diffusion-based conditional synthesis for stable high-fidelity generation, and lesion-centric augmentation via lesion generation and composition; we review these three streams and their limitations in fully synthetic 3D FLAIR generation with subtle MS lesions.

\subsection{GAN-based medical image synthesis and mask-conditioned translation}
GANs introduced adversarial learning for realistic image generation and have been widely adopted in biomedical image analysis \cite{NIPS2014_Goodfellow}. For conditional synthesis, Pix2Pix demonstrated a general-purpose framework for image-to-image translation with paired supervision, which has inspired volumetric (3D) variants for MRI synthesis using segmentation masks or anatomical labels as conditions \cite{isola2017image}. When paired data are limited, unpaired translation methods such as DiscoGAN provide another paradigm for learning cross-domain mappings \cite{kim2017learning}. DiscoGAN belongs to a broader class of cycle-consistent adversarial models for image-to-image translation, which also includes CycleGAN \cite{zhu2017cycleGAN}. Despite strong empirical performance, GAN-based pipelines can be sensitive to training dynamics and may exhibit artifacts or limited diversity, which is especially problematic in biomedical settings where small structures can be easily distorted or overlooked \cite{saad2024survey, Y.chen2022ganReview}. These observations have encouraged increasing interest in alternative generative paradigms with more stable training behavior.

\subsection{Diffusion models in medical imaging and semantic 3D brain MRI synthesis}
Denoising diffusion probabilistic models (DDPMs) 
+provide a likelihood-based generative framework that has shown strong fidelity and improved training stability compared to adversarial methods in many imaging domains \cite{ho2020denoising}. Recent surveys further highlight the rapid adoption of diffusion models across medical imaging tasks, including synthesis, reconstruction, and augmentation \cite{kazerouni2023diffusion}. Diffusion-based generation has also been explored as a data augmentation strategy for downstream medical image analysis, demonstrating improved robustness in segmentation tasks \cite{wolleb2022diffusion, islam2024lebel}. Conditional control in diffusion models has been explored through guidance and conditioning strategies that influence the denoising trajectory without adversarial training \cite{dhariwal2021diffusion, nichol2021improved}. Recent work has also explored latent diffusion formulations to reduce computational cost while preserving high-level anatomical structure in 3D brain MRI synthesis \cite{Rombach_2022_CVPR,pinaya2022brain}. In neuroimaging, Dorjsembe et al. proposed Med-DDPM, a conditional diffusion approach for semantic 3D brain MRI synthesis, demonstrating that diffusion can generate coherent volumetric anatomy under semantic guidance \cite{Med-DDPM}. However, most existing semantic synthesis work---including diffusion-based approaches---primarily targets global anatomical realism and coarse structural control. Preserving tiny, sparse pathological regions (e.g., small MS lesions) remains challenging because such regions contribute minimally to global image losses and can be smoothed during iterative denoising, especially under severe lesion--background imbalance.

\subsection{Lesion-centric generation and compositing-based augmentation}
A complementary line of research focuses on lesion-centric augmentation, where lesions are generated and then inserted into real images to enhance downstream segmentation. For example, Huo et al. studied self-supervised lesion generation and composition to produce augmented training samples \cite{Huo2024LesionGen}. While effective for improving segmentation robustness, such composition-based strategies preserve real background voxels and thus produce composited images rather than end-to-end generated 3D intensity volumes. The difficulty of preserving small pathological regions in these settings is closely related to severe lesion--background imbalance, which has motivated region-aware and class-balanced objectives in medical image analysis \cite{sudre2017generalised}. As a result, fully generative 3D MRI synthesis that simultaneously maintains whole-brain anatomical realism while explicitly prioritizing subtle lesion fidelity remains an open challenge.
\section{Methods and Experiments}
\label{sec:02}

\subsection{Model Architecture}
We formulate Lesion-DDPM as a conditional Gaussian diffusion model for 3D FLAIR MRI synthesis. The target volume is a preprocessed FLAIR MRI scan in which skull-stripped, bias-corrected brain tissue and multiple-sclerosis lesions appear as hyperintense voxels against normal-appearing white and gray matter, and each volume is resampled to a fixed resolution and intensity range for 3D convolution. Given a clean target volume $x_0 \in {R}^{1 \times D \times H \times W}$ (channel $\times$ depth $\times$ height $\times$ width) and an anatomical mask $c$ that encodes brain and lesion regions, the forward process gradually adds Gaussian noise $\epsilon \sim \mathcal{N}(0, I)$ according to a cosine variance schedule, producing a sequence of noisy volumes $\{x_t\}_{t=1}^T$. At each diffusion step $t$, the reverse transition $p_\theta(x_{t-1} \mid x_t, c)$ is parameterized by a 3D U-Net denoiser that predicts the noise field $\epsilon_\theta(x_t, t, c)$ \cite{ho2020denoising}. A sinusoidal timestep embedding is fed into every residual block, conditioning predictions on the current noise level, while the anatomical mask $c$ provides semantic guidance on brain and lesion locations.

\begin{figure}[t]
  \centering
  \includegraphics[width=0.95\columnwidth]{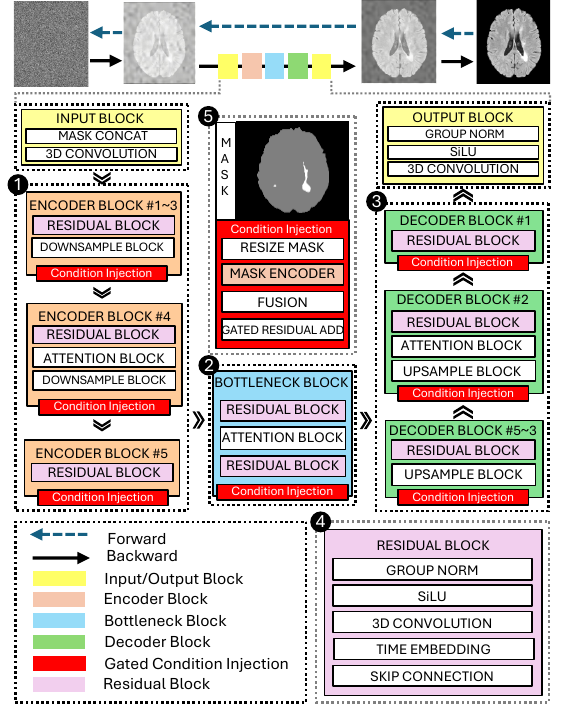}
  \caption{Overview of Lesion-DDPM.
  }
  \label{fig:my_pdf_diagram}
  \vspace{-1em}
\end{figure}


The denoiser adopts a 3D U-Net  \cite{cciccek20163d} architecture with skip connections between symmetric encoder and decoder stages as shown in Fig.~\ref{fig:my_pdf_diagram}. The encoder \blackcircle{1} consists of multiple resolution levels; each level contains one or more residual blocks and is followed by down-sampling. An attention block is inserted at an intermediate resolution to capture non-local context. The bottleneck \blackcircle{2} stacks residual and attention layers to aggregate global information. The decoder \blackcircle{3} mirrors the encoder with up-sampling and residual blocks, progressively fusing encoder features through skip connections to recover fine anatomical details. Each residual block \blackcircle{4} follows a standard design with Group Normalization, SiLU activations, 3D convolutions, timestep embedding, and a skip connection. The final output head maps the resulting feature map back to a single-channel noise prediction. This multi-scale design allows the model to capture both coarse brain geometry and fine lesion boundaries, so that the predicted noise field respects realistic tissue contrast and lesion appearance.

Beyond an optional input-level concatenation of the noisy image $x_t$ and the anatomical mask $c$, Lesion-DDPM introduces \emph{multi-level gated condition injection} \blackcircle{5}. At each encoder stage, the bottleneck, and each decoder stage, the mask is resized and encoded into a scale-matched feature map, then fused with the main stream via channel-wise concatenation followed by a pointwise 3D convolution. The fused update is added back to the main stream as a residual modulated by a learnable sigmoid gate.
Using distinct conditioning paths across depths lets lesion cues inform both global and local representations while preserving overall anatomy. Training uses the standard diffusion noise-prediction objective with a \emph{lesion-weighted} $L1$ loss that up-weights lesion voxels relative to tissue and background, improving preservation of small, sparsely distributed MS lesions.


For each encoder level, for the bottleneck, and for each decoder level, a lightweight convolutional tower first encodes the mask into a feature map at the corresponding spatial resolution. This conditional feature is then fused into the main feature stream by channel-wise concatenation followed by a point-wise 3D convolution and a learnable sigmoid gate that scales the residual update before it is added back to the original feature map. Separate condition encoders and gates are used at encoder, bottleneck, and decoder stages, enabling lesion information to influence both high-level and low-level representations while preserving global brain anatomy. In practice, the convolutional towers transform the sparse, categorical mask into scale-appropriate feature maps, the point-wise convolution learns how to mix anatomical cues with image features at each voxel, and the gates control how strongly lesion information is injected at each depth. This explicit but gated conditioning helps the network repeatedly reinforce lesion locations throughout the U-Net while avoiding over-conditioning, which we find improves preservation of small multiple-sclerosis lesions in the synthesized FLAIR images.

\subsection{Experimental Setup}
\label{sec:2.2}
We utilized the MSLesSeg dataset \cite{guarnera2025mslesseg}, which consists of 3D FLAIR MRI scans from patients with Multiple Sclerosis. The dataset provides preprocessed images with the skull already removed to focus on brain tissue. In addition, it includes high-quality lesion masks that were expert-validated as ground truth. 
From the original 115 pairs of FLAIR images and lesion masks, we retained 100 quality-controlled pairs for this study. We split the data into 80 pairs for training Lesion-DDPM and 20 pairs for testing. The native matrix size was 182 × 218 × 182 voxels. To obtain a cubic field of view and a consistent input tensor size without additional interpolation, we center-cropped each volume to 182 × 182 × 182 voxels.



We optimized the models using the Adam optimizer with an exponential moving average of 0.995 under a two-stage training schedule. Stage 1 was trained for 15,000 steps with a learning rate of $1\times10^{-5}$, and stage 2 for 10,000 steps with $1\times10^{-6}$. 
Training employed conditional mask inputs through input-level concatenation and multi-level condition injection, following a diffusion process of 250 steps. The batch size was 1 with gradient accumulation of 2 to stabilize updates. To mitigate the class imbalance described in Section \ref{sec:introduction}, stage-2 training used a lesion-weighted L1 noise-prediction loss. The lesion term was assigned substantially higher weight to counter severe lesion--background voxel imbalance and increase gradient emphasis on lesion voxels.


To provide a fair and comprehensive comparison, we compared Lesion-DDPM against three representative models (3D Pix2Pix, 3D DiscoGAN, Med-DDPM) where each was trained with the same data splits and preprocessing \cite{isola2017image, kim2017learning, Med-DDPM}. Pix2Pix and DiscoGAN were used as conditional GANs that take lesion masks as input and generate MR images.
Med-DDPM served as a diffusion-based baseline with an L1 noise-prediction objective over 250 denoising steps under the same data configuration. All models were trained for 25{,}000 steps. We fixed the total training steps for all methods to match the training budget and ensure a fair comparison. For the baselines, we used a fixed learning rate of $1\times10^{-5}$, while Lesion-DDPM adopted a two-stage schedule (15{,}000 steps at $1\times10^{-5}$ followed by 10{,}000 steps at $1\times10^{-6}$).

We further performed ablations on multi-level gated condition injection and the lesion-weighted loss: \textit{Injection-only} keeps multi-level injection but uses the standard L1 noise-prediction objective, while \textit{Weighted-only} uses only input-level mask concatenation but retains the lesion-weighted loss. All variants follow the same training setup described above.
\section{results}
\label{sec:03}

\begin{figure}[t]
  \centering
  \includegraphics[width=\linewidth]{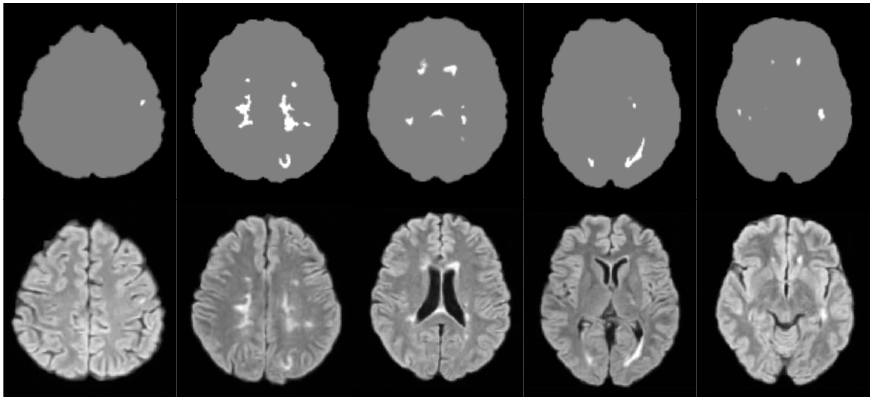}
  \caption{Input lesion masks and corresponding Lesion-DDPM FLAIR slices at different locations within one MRI volume.}
  \label{fig:lesion-ddpm_visual}
\end{figure}

Fig.~\ref{fig:lesion-ddpm_visual} illustrates input lesion masks and the corresponding synthetic FLAIR slices generated by Lesion-DDPM at different axial locations within a single MRI volume. The top row shows brain and lesion masks with varying lesion load and spatial distribution, and the bottom row shows the corresponding synthesized images. Across slices, Lesion-DDPM produces anatomically plausible brains with hyperintense lesions that closely follow the input masks, while maintaining consistent gray–white matter contrast.

\begin{figure}[t]
  \centering
  \includegraphics[width=\linewidth]{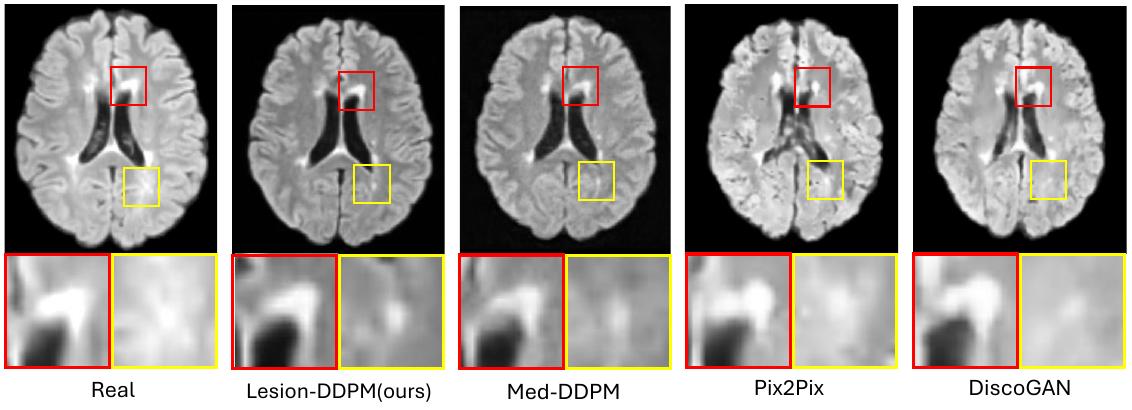}
  \caption{Visual comparison between real and synthetic MRIs}
  \label{fig:visual_comparison}
\end{figure}

Fig.~\ref{fig:visual_comparison} presents a side-by-side comparison of real FLAIR, Lesion-DDPM, and the three baselines given the same input mask. Zoomed regions (bottom row) highlight that Lesion-DDPM more faithfully preserves lesion boundaries and periventricular structures, whereas Med-DDPM tends to produce smoother but blurrier lesions and cortical textures, and the GAN-based models exhibit noisier textures and shape distortions around lesion areas.


Building on these qualitative observations, we organize the remainder of this section into three parts: we first report an expert assessment of visual realism, then present quantitative voxel-wise metrics of generation fidelity, and finally evaluate how synthetic images from each model support downstream lesion segmentation.

\subsection{Expert Assessment}

To complement qualitative comparisons with human judgment, we first conducted an expert assessment of image realism. Two neurologists were asked to select the 8 real images among 16 FLAIR images: 8 real and 8 synthetic images (2 from each generative model -  Lesion-DDPM, Med-DDPM, 3D Pix2Pix, and 3D DiscoGAN). In the final selections, the 8 images identified as real included both synthetic images generated by Lesion-DDPM. This preliminary result indicates that Lesion-DDPM can produce synthetic FLAIR images that are visually convincing enough to be judged as real by neurologists.

\subsection{Generation Performance}
To quantitatively evaluate the fidelity of synthesized 3D FLAIR volumes, we assessed voxel-level intensity and structural similarities between generated and real MRIs using three complementary metrics: Mean Squared Error (MSE), Mean Absolute Error (MAE), and Multi-Scale Structural Similarity (MS-SSIM). MSE and MAE measure voxel-wise intensity deviations, with lower values indicating closer alignment in gray-level distribution, whereas MS-SSIM reflects perceptual and structural consistency across multiple spatial scales in 3D space.

\begin{table}[h]
\caption{Quantitative evaluation on whole and lesion regions.}
\centering
\resizebox{\columnwidth}{!}{
\scriptsize
\begin{tabular}{lccc}
\hline
\textbf{Model} & \textbf{MSE $\downarrow$} & \textbf{MAE $\downarrow$} & \textbf{MS-SSIM $\uparrow$} \\
\hline
\multicolumn{4}{c}{\textit{Whole Region Metrics}} \\
\hline
\textbf{Lesion-DDPM} & \textbf{0.059 $\pm$ 0.034} & \textbf{0.202 $\pm$ 0.070} & \textbf{0.806 $\pm$ 0.045} \\
3D DiscoGAN & 0.087 $\pm$ 0.029 & 0.265 $\pm$ 0.051 & 0.790 $\pm$ 0.055 \\
3D Pix2Pix & 0.090 $\pm$ 0.030 & 0.268 $\pm$ 0.052 & 0.786 $\pm$ 0.057 \\
Med-DDPM & 0.198 $\pm$ 0.063 & 0.421 $\pm$ 0.080 & 0.728 $\pm$ 0.039 \\
\hline
\multicolumn{4}{c}{\textit{Lesion Region Metrics}} \\
\hline
\textbf{Lesion-DDPM} & \textbf{0.048 $\pm$ 0.048} & \textbf{0.181 $\pm$ 0.102} & \textbf{0.798 $\pm$ 0.039} \\
3D DiscoGAN & 0.121 $\pm$ 0.061 & 0.327 $\pm$ 0.091 & 0.740 $\pm$ 0.058 \\
3D Pix2Pix & 0.119 $\pm$ 0.060 & 0.324 $\pm$ 0.090 & 0.733 $\pm$ 0.061 \\
Med-DDPM & 0.105 $\pm$ 0.060 & 0.299 $\pm$ 0.098 & 0.733 $\pm$ 0.038 \\
\hline
\end{tabular}
}
\label{tab:metrics_mse}
\end{table}

\begin{table*}[h]
\caption{Segmentation performance comparison across real, synthetic-only, and hybrid training sets.}
\centering
\begin{tabular}{l@{\hspace{7em}}c@{\hspace{7em}}c@{\hspace{7em}}c@{\hspace{7em}}c}
\hline
\textbf{Model} & \textbf{Dice} & \textbf{IoU} & \textbf{Precision} & \textbf{Recall} \\
\hline
\multicolumn{5}{c}{\textit{Real Images Only}} \\
\hline
Real images & 0.670 ± 0.156 & 0.521 ± 0.156 & 0.697 ± 0.226 & 0.689 ± 0.129 \\
\hline
\multicolumn{5}{c}{\textit{Synthetic Images Only}} \\
\hline
3D Pix2Pix       & 0.543 ± 0.169 & 0.389 ± 0.154 & 0.544 ± 0.216 & \textbf{0.653 ± 0.202} \\
3D DiscoGAN      & 0.528 ± 0.191 & 0.379 ± 0.170 & 0.522 ± 0.225 & 0.634 ± 0.213 \\
Med-DDPM      & 0.569 ± 0.163 & 0.414 ± 0.149 & 0.556 ± 0.226 & 0.642 ± 0.125 \\
\textbf{Lesion-DDPM}   & \textbf{0.616 ± 0.139} & \textbf{0.459 ± 0.143} & \textbf{0.620 ± 0.202} & 0.651 ± 0.121 \\
\hline
\multicolumn{5}{c}{\textit{Hybrid (Real + Synthetic Images)}} \\
\hline
3D Pix2Pix       & 0.575 ± 0.166 & 0.422 ± 0.165 & 0.470 ± 0.175 & 0.778 ± 0.172 \\
3D DiscoGAN      & 0.543 ± 0.175 & 0.392 ± 0.165 & 0.432 ± 0.179 & \textbf{0.781 ± 0.170} \\
Med-DDPM      & 0.667 ± 0.153 & 0.519 ± 0.164 & 0.698 ± 0.159 & 0.663 ± 0.181 \\
\textbf{Lesion-DDPM}   & \textbf{0.685 ± 0.129} & \textbf{0.534 ± 0.145} & \textbf{0.704 ± 0.148} & 0.691 ± 0.156 \\
\hline

\end{tabular}
\label{tab:segmentation_table}
\end{table*}

\begin{figure*}[h]
  \centering
  \includegraphics[width=\textwidth]{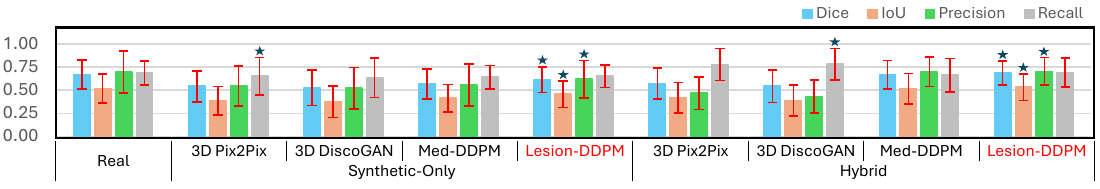}
  \caption{Downstream segmentation (Dice, IoU, Precision, Recall) on the real MRI test set across Real, Synthetic-only, and Hybrid training. Bars: mean$\pm$SD; stars mark the best per metric.}
  \label{fig:Downstream_results}
\end{figure*}

As summarized in Table~\ref{tab:metrics_mse}, where downward arrows (↓) indicate metrics for which lower values are better and the upward arrow (↑) indicates metrics for which higher values are better, our Lesion-DDPM achieved the lowest MSE (0.059) and MAE (0.202), as well as the highest MS-SSIM (0.806) among all models. These results indicate that Lesion-DDPM produces synthetic FLAIR volumes with superior voxel-level fidelity and more realistic 3D structural patterns compared to other baseline models. This advantage is consistent with the model design: the multi-level gated condition injection supplies anatomical masks to the encoder, bottleneck, and decoder blocks, which stabilizes conditioning and encourages voxel intensities to follow the underlying brain and lesion structures across scales. In addition, the lesion-weighted L1 reconstruction loss places higher penalties on errors within lesion voxels while still constraining global deviations, discouraging over-smoothing and helping the model reconstruct sharper tissue boundaries and tissue contrast. The improvement in MS-SSIM further suggests that the model effectively preserves fine-grained anatomical structures and tissue contrast while mitigating blurring and noise artifacts. In addition, we separately evaluated generation quality within lesion regions. Lesion-DDPM again demonstrated clear superiority, yielding the lowest lesion-wise MSE (0.048) and MAE (0.181), along with the highest MS-SSIM (0.798). 

Beyond voxel-level similarity, we additionally evaluated set-level MS-SSIM, following the evaluation protocol of Med-DDPM to measure structural consistency between the entire sets of synthetic and real MRIs. Lesion-DDPM achieved a mean score of 0.660, slightly outperforming 3D DiscoGAN (0.652) and 3D Pix2Pix (0.648) and more clearly surpassing Med-DDPM (0.598); the reference score for real–real pairs was 0.7112. Because set-level MS-SSIM is computed over whole collections of scans rather than individual images, Lesion-DDPM generates a synthetic cohort whose overall anatomical structures and lesion configurations are closer to those of the real dataset.

We further conduct an ablation study to assess the contribution of multi-level gated condition injection and the lesion-weighted loss, reporting results on the brain-tissue region (voxels with mask $>0$). The full Lesion-DDPM yields the highest structural similarity and the lowest reconstruction error, outperforming both Injection-only (MS-SSIM = 0.664, MSE = 0.079) and Weighted-only (MS-SSIM = 0.641, MSE = 0.074).
Notably, Weighted-only improves voxel-wise errors compared to Injection-only, while removing multi-level injection substantially degrades MS-SSIM, suggesting complementary roles of the two components. Combining both yields the best balance between voxel-level fidelity and structural consistency.

\subsection{Downstream Evaluation}

For downstream evaluation, we trained a standard 3D U-Net segmentation network using an identical training protocol for three data configurations: real-only (80 real volumes), synthetic-only (80 generated volumes), and hybrid (40 real + 40 synthetic volumes). We kept the total number of training volumes fixed (80) to control for dataset size and isolate the effect of synthetic data. All models were evaluated on the same fixed real MRI test set, and quantitative segmentation results are summarized in Table~\ref{tab:segmentation_table}.

Based on our observation, the proposed Lesion-DDPM outperformed all baseline generative models. Under the synthetic-only setting, all models exhibited a performance drop relative to the real-only baseline, as expected due to the absence of true anatomical variability. As shown in Fig.~\ref{fig:Downstream_results}, Lesion-DDPM achieved the highest Dice (0.616) and IoU (0.459), surpassing both GAN-based models and the standard Med-DDPM. Compared with the real-only reference, this synthetic-only model still retains a large fraction of segmentation accuracy, suggesting that lesions synthesized by Lesion-DDPM preserve morphology and contrast well enough for the network to learn meaningful decision boundaries from synthetic data alone.

In the hybrid configuration, segmentation accuracy improved across all models, confirming that high-quality synthetic MRI can enhance data diversity and generalization. As shown in Fig.~\ref{fig:Downstream_results}, Lesion-DDPM achieved the highest Dice (0.685) and IoU (0.534), while 3D DiscoGAN attained the highest recall (0.781). However, DiscoGAN’s higher recall was accompanied by lower precision, reflecting a tendency to over-segment lesions, whereas Lesion-DDPM maintained a more favorable balance between precision and recall. Taken together, these results indicate that Lesion-DDPM provides the most balanced segmentation performance across Dice, IoU, precision, and recall, and that its synthetic volumes not only resemble real MRIs visually but also encode lesion patterns that transfer effectively to downstream segmentation.
\section{Conclusion}
\label{sec:05}

In this study, we presented Lesion-DDPM, a 3D conditional diffusion framework specifically designed for lesion-aware FLAIR MRI synthesis in multiple sclerosis. By incorporating multi-level conditional injection and a lesion-weighted L1 objective, our method effectively preserved small lesion structures while maintaining global anatomical consistency. Expert assessment demonstrated that synthetic images generated by Lesion-DDPM were often perceived as real by neurologists, indicating high visual realism and diagnostic plausibility. Quantitative evaluations further showed that Lesion-DDPM outperformed existing GAN-based and diffusion-based baselines across MSE, MAE, and MS-SSIM metrics, while downstream segmentation experiments confirmed its utility in enhancing performance through hybrid training with real data. These findings highlight the model’s potential for data augmentation and lesion-preserving MRI synthesis under limited-data conditions. Future extensions may explore multi-modal MRI synthesis, such as incorporating T2-weighted or cross-sequence generation to enhance anatomical consistency and cross-modality generalization, as well as controllable lesion progression simulation to enable realistic modeling of lesion evolution for longitudinal analysis. 

\balance
\bibliographystyle{IEEEtran}
\bibliography{refs}

\end{document}